\documentclass[11pt]{article}

\usepackage[final]{acl}
\usepackage[most]{tcolorbox}
\usepackage{times}
\usepackage{latexsym}
\usepackage[T1]{fontenc}
\usepackage[utf8]{inputenc}
\usepackage{microtype}
\usepackage{hyperref}
\usepackage{url}
\usepackage{booktabs}
\usepackage{xspace}
\usepackage{graphicx}
\usepackage{textcomp}
\usepackage{multicol}
\usepackage{multirow}
\usepackage{enumitem}
\usepackage{pifont}
\usepackage{needspace}

\usepackage{amsmath}
\usepackage{float}
\usepackage{tikz}
\usepackage{subcaption}
\usepackage{lineno}
\usepackage{makecell}
\usepackage{CJKutf8}
\usepackage{inconsolata}
\usepackage{adjustbox}
\usepackage{placeins}
\usepackage{tabularx}
\usepackage{array}

\usepackage{placeins}
\usepackage{flafter}
\usepackage{caption}

\definecolor{darkblue}{rgb}{0, 0, 0.5}
\hypersetup{colorlinks=true, citecolor=darkblue, linkcolor=darkblue, urlcolor=darkblue}

\definecolor{mycolor}{RGB}{33, 95, 154}
\definecolor{custom_red}{RGB}{228, 54, 54}
\definecolor{darkblue}{rgb}{0, 0, 0.5}
\definecolor{rahighlight}{RGB}{235, 244, 255}

\title{Cultural Divergence Preservation: Diagnosing Flattening and Caricature in LLM-Simulated Survey Populations}

\author{
  Yeeun Chae$^{1,2}$\thanks{\ Equal contribution.} \quad
  Yewon Choi$^{1}$\footnotemark[1] \quad
  Seunghyun Lee$^{1,3}$ \quad
  IL Im$^{1}$\thanks{\ Corresponding author.} \\
  $^{1}$Yonsei University \quad $^{2}$NAVER \quad $^{3}$Seoul National University Hospital \\
  \texttt{\{chynn2, yewon0126, lutris, il.im\}@yonsei.ac.kr}
}

\begin{document}
\maketitle

\begin{abstract}
Large language models (LLMs) are increasingly used as synthetic survey respondents to estimate population response distributions.
In cross-cultural survey simulation, evaluations should assess not only distributional fidelity within countries but also whether differences across countries are preserved.
However, existing distance-based metrics such as Jensen--Shannon divergence (JSD) do not directly capture such cross-country differences.
To address this limitation, we introduce Cultural Divergence Preservation (CDP), a reference-light diagnostic based on a one-time human calibration.
CDP identifies reduced cross-country divergence as cultural flattening and increased divergence as cultural caricature.
To evaluate CDP, we conduct experiments across four LLM backbones, three persona-based prompting methods, and two survey domains, the World Values Survey (WVS) and the Big Five Personality Test.
The results reveal a systematic discrepancy between conventional fidelity metrics and CDP.
Controlled experiments show that CDP changes monotonically as cross-country divergence is attenuated or amplified, while the corresponding changes in JSD remain relatively small.
In our audit of real LLM generations, DeepPersona-Inspired prompting is frequently favored by conventional fidelity metrics but exhibits the strongest flattening in every model--domain block.
CDP thus complements fidelity metrics by directly quantifying the attenuation or amplification of cross-country divergence.

\end{abstract}

\begin{figure}
\centering
\includegraphics[width=\columnwidth]{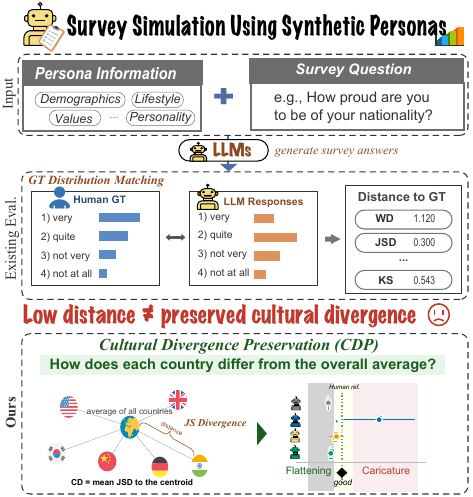}
\caption{Motivations of CDP.}
\label{fig:concept}
\end{figure}

\section{Introduction}
\label{sec:intro}

Large language models (LLMs) are increasingly used as
\emph{synthetic survey respondents} for estimating population response
distributions
\citep{park2022social, aher2023using, argyle2023out, santurkar2023whose,
chen2025specializing}. This enables low-cost pilot studies, larger synthetic
samples, and broader coverage of under-surveyed populations.

Many survey applications require both country-level distributional fidelity
and the preservation of differences across populations. Existing evaluations compare synthetic
and human response distributions using various aggregate distributional
metrics, including Wasserstein distance (WD), Jensen--Shannon divergence
(JSD), and related variants
\citep{santurkar2023whose, durmus2023towards, wang2025deeppersona}. While
effective for measuring
overall distributional error, these metrics do not explicitly evaluate whether
cross-country variation---and thus one measurable aspect of cultural
pluralism---is preserved. As a result, synthetic populations may suppress
cross-country differences (flattening) or exaggerate them (caricature) while
receiving similar distribution-matching scores.

To address this limitation, we propose \textbf{Cultural Divergence Preservation
(CDP)}, a reference-light diagnostic that quantifies cross-country response divergence preservation relative to a human reference. CDP distinguishes whether cultural pluralism is flattened, preserved, or exaggerated, while requiring only a one-time human calibration for each survey domain that can be reused across models and prompting methods. We evaluate CDP through a controlled experiment that systematically attenuates and amplifies cross-country divergence, and an empirical audit spanning four LLMs, three prompting methods, and two survey domains.

We make the following contributions:
(1) We introduce \textbf{CDP}, a reference-light diagnostic for evaluating cultural divergence preservation in synthetic survey data. (2) We conduct a controlled sensitivity check with systematically attenuated and amplified data sets, showing that CDP responds monotonically while the induced JSD remains small relative to the distributional error observed in real LLM conditions. (3) We audit four LLMs with three prompting methods on two survey domains and show that DeepP-I, the method most frequently favored by conventional distribution-matching metrics, exhibits the strongest flattening in all eight model--domain blocks. This result demonstrates that country-level fidelity rankings do not separately characterize divergence preservation.

\section{Related Work}

Most prior work evaluates synthetic surveys using reference-based fidelity
metrics that compare model-generated and matched human response distributions
within each population
\citep{santurkar2023whose, durmus2023towards,
alkhamissi2024investigating, ma2025algorithmic}.
Such metrics quantify population-level discrepancy but do not separately
characterize cross-population structure.

Recent studies document two opposing distortions: variance flattening
suppresses subgroup- and identity-level diversity
\citep{bisbee2024synthetic, dominguez2024questioning,
wang2025large},
whereas caricature exaggerates stereotypical differences
\citep{cheng2023compost}.
Persona conditioning does not consistently eliminate either problem
\citep{sun2025sociodemographic, hu2024quantifying}.
Detecting these distortions generally requires condition-matched human data
\citep{morocho2026assessing}.
Approaches that reduce this dependence instead assess textual or other fidelity
dimensions rather than cross-population distributional structure
\citep{hullman2026, zhang2026survey, alaa2022faithful,
batzner2025whose, dash2025polypersona, choi2026beyond}.

Cross-country structure has also been evaluated through cultural profiles,
country rankings, and retention of value signals and within-country
diversity
\citep{masoud2025cultural, luther2025american, agarwal2026plural}.
These approaches capture important cultural variation but do not summarize
whether divergence among full country-level response distributions is
attenuated or amplified relative to a reusable human baseline. CDP complements
them by calibrating aggregate divergence from the cross-country centroid
against the corresponding human level.

\section{Cultural Divergence Preservation}
\label{sec:method}

\subsection{Definition}
\label{sec:method-def}

Let $C$ be a set of countries and $Q$ a set of survey items, where each item
$q \in Q$ is answered on a discrete rating scale. For country $c$ and item $q$,
let $p_{c,q}$ denote the normalized response histogram, i.e., the vector of
response-option proportions observed in the data under evaluation. For each
item, we compute the equal-weight centroid of the country histograms:
\begin{equation}
\bar{p}_{q} \;=\; \frac{1}{|C|}\sum_{c \in C} p_{c,q}.
\label{eq:centroid}
\end{equation}
We weight countries equally so that the centroid reflects cross-country
structure rather than differences in national sample size.

For two discrete probability distributions $p$ and $r$, let
$D_{\mathrm{JS}}(p \,\|\, r)$ denote their Jensen--Shannon divergence
\citep{lin1991divergence}, computed with base-2 logarithms so that it is
bounded in $[0,1]$; it is zero when the two distributions are identical and
increases as they become more dissimilar.

We define \textbf{Cultural Divergence (CD)} as the mean JS divergence between each country's response
distribution and the corresponding cross-country centroid:
\begin{equation}
\mathrm{CD}
\;=\; \frac{1}{|Q|\,|C|}
\sum_{q \in Q}\sum_{c \in C}
D_{\mathrm{JS}}\!\left(p_{c,q} \,\middle\|\, \bar{p}_{q}\right).
\label{eq:cd}
\end{equation}
CD is zero when all countries share the same response distribution and
increases as cross-country response distributions become more divergent.

CD operationalizes the \emph{magnitude} of cross-country response divergence,
which cross-cultural research treats as substantively meaningful
\citep{hofstede2011dimensionalizing, inglehart2005modernization}.

\subsection{Human Reference}
\label{sec:method-ref}

Because the natural magnitude of cross-country divergence varies across
domains, we express synthetic CD relative to a domain-specific human reference
and define
\textbf{Cultural Divergence Preservation (CDP)} as
$\mathrm{CDP}=\mathrm{CD}_{\mathrm{synthetic}}/\mathrm{CD}_{\mathrm{human}}$,
reported as a percentage. Values below 100\% indicate attenuated divergence
(flattening), and values above 100\% indicate amplified divergence
(caricature).

Exact country-wise matching entails CDP $=100\%$. With nonzero generation
error, however, average country-wise fidelity does not determine whether
cross-country divergence is attenuated or amplified.

Although CD and conventional fidelity JSD use the same divergence function,
they compare different distribution pairs: fidelity JSD compares each
synthetic country distribution with its human counterpart, whereas CD compares
each country distribution with the cross-country centroid. Thus, fidelity JSD
quantifies country-wise distributional error, while CDP summarizes the
preservation of between-country divergence.

Human CD is estimated once per domain and reused across synthetic conditions,
but must be recomputed when the country set, item set, or response scale changes. Domain-level human CD values are reported in
Appendix~\ref{app:flattening-details}, with per-country values in
Appendix~\ref{app:country-cdp}.

\section{Experiments and Results}
\label{sec:experiment}
To validate the CD measure developed in this study, we conduct several experiments.
The experiments compare persona-based data synthesis methods to verify whether CDP captures how cultural diversity is preserved in the synthetic data.
%%예원작성

\subsection{Experimental Setup}
\label{sec:setup}

\paragraph{Datasets.}
We adopt the two survey instruments used in DeepPersona \citep{wang2025deeppersona}---the World Values Survey and the Big Five Personality Test---while following the same country selection and questionnaire items.
For the social-values audit, we use six WVS Wave~7 items \citep{haerpfer2022world} and national response distributions for Argentina, Australia, Germany, India, Kenya, and the United States.
%\footnote{\url{https://www.worldvaluessurvey.org/WVSDocumentationWV7.jsp}}
For the personality audit, we use the 50-item IPIP Big Five inventory \citep{goldberg1992development} for Argentina, Australia, and India, with human-response distributions from OpenPsychometrics.
% \footnote{\url{https://ipip.ori.org/new_ipip-50-item-scale.htm}}
%\footnote{\url{https://openpsychometrics.org/tests/IPIP-BFFM/}}
For the real-generation audit, these country-level human response distributions serve solely as evaluation references; the controlled sensitivity analysis uses them as the starting distributions for perturbation. Country-level sample sizes are reported in Appendix~\ref{app:sample-sizes}.
All survey items are provided in Appendix~\ref{app:items}.

\paragraph{Synthetic-Persona-based survey simulation.}
We generate 300 simulated respondents for each country, persona method, and survey model.
We compare three persona-construction methods: Cultural Prompting \citep{tao2024cultural}, PersonaHub-Inspired (PHub-I), and DeepPersona-Inspired (DeepP-I).
\textbf{Cultural Prompting} specifies only that the respondent was born in and currently lives in the target country.
\textbf{PHub-I} samples one-line personas from PersonaHub \citep{ge2024scaling}, assigns them to the target country, and expands them into structured profiles using the OpenCharacter prompt \citep{wang2025opencharacter}.
\textbf{DeepP-I} samples demographic, personality, belief, social, and lifestyle attributes from a fixed taxonomy and renders them using a predefined natural-language template following DeepPersona.
We conduct the survey simulations using four open-weight models: Gemma-3-4B \citep{gemmateam2025gemma3}, Qwen3.5-9B and Qwen3.5-27B \citep{team2026qwen3}, and Llama-2-13B \citep{touvron2023llama2}.
Detailed prompts and persona-construction procedures are provided in Appendix~\ref{sec:prompts}.

\paragraph{Evaluation.}
We measure country-level distributional fidelity using WD, JSD, and KS, which are widely used in prior work on survey simulation and response-distribution modeling \citep{wang2025deeppersona,chen2025rose}.
As a complementary diagnostic, we evaluate CDP by comparing synthetic CD with that observed in the corresponding human survey data (\S\ref{sec:method}).
%%%%/예원작성

\subsection{Controlled Flattening and Exaggeration}
\label{sec:exp1}

Starting from the human distributions, we scale each country--item deviation
from the centroid as
\[
p_{c,q}^{(\gamma)}=\bar{p}_{q}+\gamma(p_{c,q}-\bar{p}_{q}),
\qquad \gamma\geq0.
\]
Values $0\leq\gamma<1$ move countries toward the centroid (equivalently,
$\lambda=1-\gamma$), with complete collapse at $\gamma=0$; $\gamma=1$
recovers the human data; and $\gamma>1$ amplifies country deviations. For
$\gamma>1$, negative bins are clipped to zero and histograms are renormalized
(Appendix~\ref{app:flattening-details}).
For $\gamma\leq1$, the transformation preserves each item-level centroid.

\begin{figure}[t]
\centering
\includegraphics[width=\columnwidth]{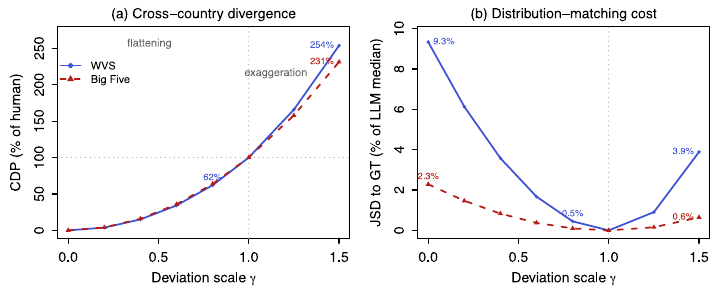}
\caption{Controlled flattening ($\gamma<1$, equivalently $\lambda=1-\gamma$)
and exaggeration ($\gamma>1$) of the human data.
(a) CDP relative to the unmodified human CD at $\gamma=1$.
(b) Induced JSD cost relative to the median JSD observed across real LLM
conditions in the same domain.}
\label{fig:controlled_flattening}
\end{figure}

% AUDIT-CORRECTED [C3]: corrected WVS-to-human distances and JSD denominator.
Figure~\ref{fig:controlled_flattening} normalizes CD by human CD and induced
JSD by the same-domain median across real LLM conditions. CDP decreases
monotonically under flattening: on WVS, 20\% centroid mixing leaves 62\% of
human divergence while incurring only 0.5\% of the LLM-median JSD. Complete
collapse makes the induced JSD equal to human CD, yet this amounts to only
9.3\% and 2.3\% of the LLM median on WVS and Big Five, respectively. Under
exaggeration, $\gamma=1.5$ raises CDP to 231--254\% while
costing only 0.6--3.9\% of the LLM-median JSD. This monotone bidirectional
response provides a controlled sensitivity check for CDP. Full results,
including WD and KS, appear in Appendix~\ref{app:flattening-details}.

\subsection{Country-Level Fidelity and Divergence Preservation}
\label{sec:exp2}

% AUDIT-CORRECTED [C4]: corrected WVS coding; nWD controls unequal item scales.
Table~\ref{tab:distribution-cdp-results} compares the three
country-conditioned prompting methods across eight model--domain blocks.
Because WVS response scales vary, nWD divides item-level WD by the response
range before averaging. DeepP-I attains the lowest raw WD in seven of
eight blocks and the lowest nWD and JSD in six, but also shows the strongest
flattening in every block (CDP: 14--59\%). Cultural is over-divergent
(184--501\%), while PHub-I is intermediate (44--99\%). These contrasting
rankings show that country-level fidelity does not separately characterize
divergence preservation.

We assess uncertainty in the human--synthetic CD gap using a respondent-level
bootstrap ($B{=}2{,}000$). The DeepP-I gap excludes zero in seven of eight
model--domain blocks, with Llama-2-13B on Big Five as the sole exception, while
all eight gaps remain positive under leave-one-country-out analysis. Full
uncertainty and robustness analyses appear in Appendix~\ref{app:robustness}.

% Main distribution-matching and CDP results; domain-specific tables are retained as alternatives.
\begin{table}[!t]
\centering
\small
\setlength{\tabcolsep}{1.5pt}
\renewcommand{\arraystretch}{0.92}
\begin{adjustbox}{max width=\columnwidth}
\begin{tabular}{lllrrrr}
\toprule
Domain & Model & Method & WD$\downarrow$ & nWD$\downarrow$ & JSD$\downarrow$ & CDP \\
\midrule
\multirow{12}{*}{WVS} & \multirow{3}{*}{Gemma-3-4B} & Cultural & 1.072 & 0.322 & 0.402 &  289\% \\
 &  & PHub-I & \textbf{0.953} & \textbf{0.312} & \textbf{0.331} &  \textbf{99\%} \\
 &  & DeepP-I & 1.006 & 0.324 & 0.339 &  16\%$^{\dagger}$ \\
\cmidrule(lr){2-7}
 & \multirow{3}{*}{Qwen3.5-9B} & Cultural & 0.816 & 0.228 & 0.297 &  461\% \\
 &  & PHub-I & 0.802 & 0.255 & 0.181 &  \textbf{68\%}$^{\dagger}$ \\
 &  & DeepP-I & \textbf{0.497} & \textbf{0.165} & \textbf{0.133} &  28\%$^{\dagger}$ \\
\cmidrule(lr){2-7}
 & \multirow{3}{*}{Qwen3.5-27B} & Cultural & 0.762 & 0.232 & 0.247 &  439\% \\
 &  & PHub-I & 0.665 & 0.222 & 0.123 &  \textbf{61\%}$^{\dagger}$ \\
 &  & DeepP-I & \textbf{0.486} & \textbf{0.163} & \textbf{0.085} &  27\%$^{\dagger}$ \\
\cmidrule(lr){2-7}
 & \multirow{3}{*}{Llama-2-13B} & Cultural & 0.825 & 0.266 & 0.251 &  184\% \\
 &  & PHub-I & 0.780 & \textbf{0.237} & \textbf{0.189} &  \textbf{44\%}$^{\dagger}$ \\
 &  & DeepP-I & \textbf{0.668} & 0.243 & 0.224 &  14\%$^{\dagger}$ \\
\midrule
\multirow{12}{*}{\shortstack{Big\\Five}} & \multirow{3}{*}{Gemma-3-4B} & Cultural & 1.004 & 0.251 & 0.487 &  389\% \\
 &  & PHub-I & 0.676 & 0.169 & 0.199 &  \textbf{65\%} \\
 &  & DeepP-I & \textbf{0.672} & \textbf{0.168} & \textbf{0.175} &  32\%$^{\dagger}$ \\
\cmidrule(lr){2-7}
 & \multirow{3}{*}{Qwen3.5-9B} & Cultural & 0.928 & 0.232 & 0.387 &  384\% \\
 &  & PHub-I & 0.842 & 0.211 & 0.196 &  \textbf{54\%}$^{\dagger}$ \\
 &  & DeepP-I & \textbf{0.475} & \textbf{0.119} & \textbf{0.123} &  35\%$^{\dagger}$ \\
\cmidrule(lr){2-7}
 & \multirow{3}{*}{Qwen3.5-27B} & Cultural & 0.883 & 0.221 & 0.343 &  501\% \\
 &  & PHub-I & 0.837 & 0.209 & 0.173 &  \textbf{48\%}$^{\dagger}$ \\
 &  & DeepP-I & \textbf{0.575} & \textbf{0.144} & \textbf{0.140} &  31\%$^{\dagger}$ \\
\cmidrule(lr){2-7}
 & \multirow{3}{*}{Llama-2-13B} & Cultural & 0.892 & 0.223 & 0.386 &  265\% \\
 &  & PHub-I & 0.699 & 0.175 & 0.242 &  \textbf{81\%} \\
 &  & DeepP-I & \textbf{0.629} & \textbf{0.157} & \textbf{0.191} &  59\% \\
\bottomrule
\end{tabular}
\end{adjustbox}
\caption{Results by domain, model, and prompt. Lower WD/nWD/JSD is better; $\mathrm{CDP}=100\%$ matches human divergence. Bold: best distance or CDP closest to $100\%$ per block; $^{\dagger}$: 95\% bootstrap CI for the human--synthetic CD gap excludes zero.}
\label{tab:distribution-cdp-results}
\end{table}

\section{Conclusion}
\label{sec:conclusion}

This work distinguishes country-level distributional fidelity from the
preservation of cross-country divergence. Although exact country-wise matching
entails both, the two evaluations can yield different rankings under generation
error. Controlled perturbations show that CDP tracks flattening and
exaggeration even when the induced JSD is small relative to observed LLM error.
Across real generations, DeepP-I is frequently favored by fidelity metrics
but is consistently the most flattened, while Cultural prompting amplifies
divergence. CDP therefore complements, rather than replaces, existing metrics
by directly diagnosing both distortions.

\section*{Limitations}

CDP has several important limitations. First, CD and CDP characterize only the
magnitude of country-associated divergence; they do not assess whether
country-specific deviations align with their human counterparts, whether the
human country ordering is recovered, or country-specific fidelity. This limitation may partly stem from CDP's reliance on JSD as its base divergence: because JSD treats response-option probabilities as unordered coordinates on a simplex, it is insensitive to which side of the centroid a country falls on, so two countries that swap positions relative to the centroid could in principle leave CD---and thus CDP---unchanged even though the underlying cross-country pattern has inverted. Detecting such inversions would require a distance that respects the ordinal structure of Likert-type scales (e.g., Wasserstein distance) or an explicit directional check such as rank correlation between synthetic and human country orderings; we leave this extension to future work. CDP should
therefore not be used as a general quality score or
downstream-fidelity predictor; it diagnoses the specific risks of flattening
and caricature. Second, although CD can be computed from synthetic responses,
CDP requires a domain-level human reference. The procedure is thus
reference-light, and the
reference must be recomputed when the countries, items, or response scales
change. Moreover, CDP does not isolate culture from other country-associated
sources of variation: survey year, translation, response styles, and sample
composition may all contribute to the observed differences. In particular, DeepP-I draws demographic and personality attributes from a fixed taxonomy that is not conditioned on the target country (\S\ref{sec:setup}), so its strong flattening may partly reflect this decontextualized sampling design rather than an inherent limitation of the underlying LLM.

The empirical scope is also limited to four open-weight models, two survey
domains, and English-language prompts. Big Five includes only three countries,
so its leave-one-country-out analysis retains only two and is best viewed as a
weak composition-sensitivity check rather than an uncertainty estimate.
Generation-seed sensitivity is limited to three seeds per condition
(Appendix~\ref{app:seed-sensitivity}). The percentile-bootstrap gap for
Llama-2-13B on Big Five includes zero, despite a positive point estimate and a
positive leave-one-country-out gap, so statistical separation is supported in
seven rather than all eight blocks. Finally, the primary WVS analysis is
unweighted. Applying the within-country \texttt{W\_WEIGHT} variable changes
aggregate human CD by only $-0.80\%$ and changes neither model-by-metric winners
nor CDP interpretation classes, but broader validation should examine weighting
choices explicitly. Future work should extend CDP across more countries,
domains, model families, languages, and generation seeds, while pairing it with
direction-sensitive and item-dependence diagnostics that evaluate which
cross-country differences are preserved and whether they support downstream
use.

\bibliography{custom}
\clearpage
\appendix
% =====================================================
% Prompt box definition (appendix-local)
% =====================================================
\newtcblisting{promptbox}[1]{
  enhanced, listing only, breakable,
  colback=white, colframe=black!100,
  boxrule=1.5pt, arc=4pt,
  left=8pt, right=8pt, top=10pt, bottom=8pt,
  listing options={
    basicstyle=\ttfamily\small,
    breaklines=true,
    breakindent=0pt,
    breakautoindent=false,
    columns=fullflexible,
    keepspaces=true,
    literate={•}{{\textbullet\ }}1,
  },
  attach boxed title to top left={xshift=10pt, yshift=-8pt},
  boxed title style={
    colback=black, colframe=black,
    arc=3pt, boxrule=0pt,
    left=6pt, right=6pt, top=2pt, bottom=2pt,
  },
  coltitle=white,
  fonttitle=\bfseries\small\sffamily,
  title={#1},
}

% =====================================================
% Appendix content
% =====================================================
\section*{Appendix}

\section{Prompts}
\label{sec:prompts}

\subsection{Cultural}
\label{app:prompt-cultural}

\begin{promptbox}{Cultural}
You are an average human being born in {country} and living in {country} responding to the following survey question. Respond with ONLY a single integer number on the given scale. Do not add any explanation or reasoning.
\end{promptbox}

\vspace{1 cm}

\subsection{PersonaHub-Inspired}
\label{app:prompt-personahub}

\begin{promptbox}{Stage 1: Character Profile Synthesis}
You are a helpful assistant. I will provide you with a short persona description. Your task is to create a character based on the given persona.

You can output a brief character description containing the following information: character name, age, gender, race, birth place, appearance, general experience, and personality.

Note:
1. Your response should start with "Name:".
2. Your character description should be specific and consistent with the persona.

{persona with appended country clause}
\end{promptbox}

\begin{promptbox}{Stage 2: Survey Elicitation}
You are the following character:

{synthesized eight-field character profile}

Answer the following survey question from this character's personal perspective, reflecting the character's background and life experience. Respond with ONLY a single integer number on the given scale. Do not add any explanation or reasoning.
\end{promptbox}

\vspace{2cm}

\subsection{DeepPersona-Inspired}
\label{app:prompt-deeppersona}

\begin{promptbox}{DeepP-I}
I am a {age}-year-old {gender} living in {area}, {country}.

Demographics:
- Education: {education level}
- Marital status: {marital status}, {children}
- Occupation: {occupation} ({career stage})
- Income: {income level}
- Housing: {housing}

Personality:
- Openness: {openness}
- Conscientiousness: {conscientiousness}
- Extraversion: {extraversion}
- Agreeableness: {agreeableness}
- Neuroticism: {neuroticism}

Values and beliefs:
- Core values: {core values}
- Life attitude: {life attitude}
- Religion: {religion} ({religiosity})
- Political leaning: {political leaning}
- Cultural identity: {cultural identity}

Social life:
- Social circle: {social circle}
- Community: {community involvement}
- Media: {media consumption}

Lifestyle:
- Health: {health status}
- Exercise: {exercise habits}
- Tech proficiency: {tech proficiency}
- Financial attitude: {financial attitude}
- Interests: {primary interest} and {secondary interest}

Answer the following survey question from my personal perspective, reflecting the background, personality, values, and life circumstances described above. Respond with ONLY a single integer number on the given scale. Do not add any explanation or reasoning.
\end{promptbox}
%\vspace{5cm}

%%%

%%%% Sec B
\clearpage
\onecolumn
\section{WVS and Big Five Dataset Details}
\label{app:dataset-details}
\subsection{Survey Items}
\label{app:items}
\begin{table}[H]
\centering
\footnotesize
\setlength{\tabcolsep}{4pt}
\renewcommand{\arraystretch}{0.95}
\begin{tabularx}{\textwidth}{
@{}
l
>{\raggedright\arraybackslash}X
>{\raggedright\arraybackslash}p{0.30\textwidth}
@{}
}
\toprule
ID & Construct and survey item & Response scale \\
\midrule
Q45
& \textbf{Respect for Authority}\par
If greater respect for authority takes place in the near future, do you think it would be a good thing, a bad thing, or you don't mind?
& 1 = A good thing; 2 = Don't mind; 3 = A bad thing \\

Q46
& \textbf{Feeling of Happiness}\par
Taking all things together, rate how happy you would say you are.
& 1 = Very happy; 2 = Quite happy; 3 = Not very happy; 4 = Not at all happy \\

Q57
& \textbf{Trust on People}\par
Generally speaking, would you say that most people can be trusted or that you need to be very careful in dealing with people?
& 1 = Most people can be trusted; 2 = Need to be very careful \\

Q184
& \textbf{Justifiability of Abortion}\par
How justifiable do you think abortion is?
& 1 = Never justifiable \dots\ 10 = Always justifiable \\

Q218
& \textbf{Petition Signing}\par
Have you signed a petition?
& 1 = Have done; 2 = Might do; 3 = Would never do \\

Q254
& \textbf{Pride of Nationality}\par
How proud are you to be your nationality?
& 1 = Very proud; 2 = Quite proud; 3 = Not very proud; 4 = Not at all proud \\
\bottomrule
\end{tabularx}
\caption{World Values Survey items.}
\label{tab:appwvs}
\end{table}

% Auto-generated (gen_paper_assets.py) from config/experiment_config.py:BF_ITEMS.
\begin{table}[H]
\centering\footnotesize
\renewcommand{\arraystretch}{0.95}
\begin{tabularx}{\linewidth}{l X l X}
\toprule
ID & Item & ID & Item \\
\midrule
EXT1 & I am the life of the party. & AGR6 & I have a soft heart. \\
EXT2 & I don't talk a lot. & AGR7 & I am not really interested in others. \\
EXT3 & I feel comfortable around people. & AGR8 & I take time out for others. \\
EXT4 & I keep in the background. & AGR9 & I feel others' emotions. \\
EXT5 & I start conversations. & AGR10 & I make people feel at ease. \\
EXT6 & I have little to say. & CSN1 & I am always prepared. \\
EXT7 & I talk to a lot of different people at parties. & CSN2 & I leave my belongings around. \\
EXT8 & I don't like to draw attention to myself. & CSN3 & I pay attention to details. \\
EXT9 & I don't mind being the center of attention. & CSN4 & I make a mess of things. \\
EXT10 & I am quiet around strangers. & CSN5 & I get chores done right away. \\
EST1 & I get stressed out easily. & CSN6 &
\resizebox{\linewidth}{!}{I often forget to put things back in their proper place.} \\
EST2 & I am relaxed most of the time. & CSN7 & I like order. \\
EST3 & I worry about things. & CSN8 & I shirk my duties. \\
EST4 & I seldom feel blue. & CSN9 & I follow a schedule. \\
EST5 & I am easily disturbed. & CSN10 & I am exacting in my work. \\
EST6 & I get upset easily. & OPN1 & I have a rich vocabulary. \\
EST7 & I change my mood a lot. & OPN2 & I have difficulty understanding abstract ideas. \\
EST8 & I have frequent mood swings. & OPN3 & I have a vivid imagination. \\
EST9 & I get irritated easily. & OPN4 & I am not interested in abstract ideas. \\
EST10 & I often feel blue. & OPN5 & I have excellent ideas. \\
AGR1 & I feel little concern for others. & OPN6 & I do not have a good imagination. \\
AGR2 & I am interested in people. & OPN7 & I am quick to understand things. \\
AGR3 & I insult people. & OPN8 & I use difficult words. \\
AGR4 & I sympathize with others' feelings. & OPN9 & I spend time reflecting on things. \\
AGR5 & I am not interested in other people's problems. & OPN10 & I am full of ideas. \\
\bottomrule
\end{tabularx}
\caption{Big Five personality items rated on a 5-point scale (1 = Disagree strongly; 5 = Agree strongly).}
\label{tab:appbf}
\end{table}

\subsection{Human Sample Sizes}
\label{app:sample-sizes}
\begin{table}[H] \centering \footnotesize \renewcommand{\arraystretch}{0.85} \setlength{\tabcolsep}{3pt} \begin{tabular*}{\textwidth}{@{\extracolsep{\fill}}cccccc@{\hspace{1.2em}}ccc@{}} \toprule \multicolumn{6}{c}{\textbf{WVS}} & \multicolumn{3}{c}{\textbf{Big Five}} \\ \cmidrule(lr){1-6} \cmidrule(lr){7-9} Argentina & Australia & Germany & India & Kenya & United States & Argentina & Australia & India \\ 1{,}003 & 1{,}813 & 1{,}528 & 1{,}692 & 1{,}266 & 2{,}596 & 486 & 8{,}584 & 2{,}820 \\ \bottomrule \end{tabular*} \caption{Country-level human sample sizes for WVS and Big Five.} \label{tab:sample-sizes}
\end{table}

%%%%%%%

\clearpage

\section{Controlled Flattening and Exaggeration Details}
\label{app:flattening-details}

Table~\ref{tab:flattening-full} reports the complete deterministic mixing grid.
For each item, the country histogram is moved toward the equal-weight country
centroid, so the manipulation is not affected by national sample sizes and
requires no simulation seed. CD falls monotonically in both domains. At
$\lambda=1$, every country distribution equals the centroid; consequently, the
induced mean JSD to the original human distributions is exactly the human CD
at $\lambda=0$. Despite eliminating all cross-country divergence, this cost is
only 9.33\% of the real-condition JSD median for WVS and 2.28\% for Big Five.
We use JSD as the primary distortion scale because it is bounded, matches the
divergence used to define CD, and yields this exact identity at complete
convergence. WD and KS are reported in Table~\ref{tab:flattening-full} as
complementary checks that the pattern is not specific to JSD. In both
controlled-manipulation tables, GT denotes the original country-level human
distribution, and JSD, WD, and KS to GT are averaged over country--item pairs.
The normalized JSD column divides JSD to GT by the same-domain median across
real LLM conditions (0.23468 for WVS; 0.20426 for Big Five).

\begin{table}[t]
\centering
\small
\setlength{\tabcolsep}{4pt}
\begin{adjustbox}{max width=\textwidth}
\begin{tabular}{llrrrrrr}
\toprule
Domain & $\lambda$ & CD & CDP (\%) & JSD$\to$GT & WD$\to$GT & KS$\to$GT & \shortstack{JSD / LLM\\median (\%)} \\
\midrule
\multirow{6}{*}{WVS} & 0.0 & 0.02190 & 100.0 & 0.00000 & 0.00000 & 0.00000 & 0.00 \\
 & 0.2 & 0.01358 & 62.0 & 0.00105 & 0.06704 & 0.02552 & 0.45 \\
 & 0.4 & 0.00749 & 34.2 & 0.00392 & 0.13407 & 0.05104 & 1.67 \\
 & 0.6 & 0.00329 & 15.0 & 0.00839 & 0.20111 & 0.07656 & 3.58 \\
 & 0.8 & 0.00082 & 3.7 & 0.01439 & 0.26814 & 0.10208 & 6.13 \\
 & 1.0 & 0.00000 & 0.0 & 0.02190 & 0.33518 & 0.12760 & 9.33 \\
\midrule
\multirow{6}{*}{Big Five} & 0.0 & 0.00467 & 100.0 & 0.00000 & 0.00000 & 0.00000 & 0.00 \\
 & 0.2 & 0.00297 & 63.7 & 0.00019 & 0.02578 & 0.01099 & 0.09 \\
 & 0.4 & 0.00167 & 35.8 & 0.00076 & 0.05157 & 0.02198 & 0.37 \\
 & 0.6 & 0.00074 & 15.9 & 0.00169 & 0.07735 & 0.03297 & 0.83 \\
 & 0.8 & 0.00019 & 4.0 & 0.00300 & 0.10314 & 0.04396 & 1.47 \\
 & 1.0 & 0.00000 & 0.0 & 0.00467 & 0.12892 & 0.05494 & 2.28 \\
\bottomrule
\end{tabular}
\end{adjustbox}
\caption{Controlled flattening. CDP is CD relative to human CD; GT is the original country distribution. JSD/LLM median scales JSD$\to$GT by the same-domain LLM median.}
\label{tab:flattening-full}
\end{table}

Table~\ref{tab:exaggeration-full} reports the affine extension in the
over-divergence direction. Each country--item histogram is transformed as
$p_{c,q}^{(\gamma)} = \bar{p}_{q} + \gamma\,(p_{c,q}-\bar{p}_{q})$; for
$\gamma\le1$ this is exactly the mixing grid above with $\lambda=1-\gamma$,
and for $\gamma>1$ negative bins are clipped to zero and the histogram is
renormalized. Because clipping perturbs the centroid, CD is recomputed
against the centroid of the transformed histograms rather than the original
one. At the maximum evaluated level, $\gamma=1.5$, clipping is limited but
nonzero: the mean clipped magnitude is at most 0.0014 per histogram and the
maximum is 0.0313. Here, clipped magnitude denotes the total magnitude of
negative entries set to zero before renormalization; both its mean and maximum
are reported per level. We stop at $\gamma=1.5$ because stronger amplification
causes non-negligible boundary clipping for some WVS histograms and is no
longer a clean affine perturbation. CDP increases monotonically in $\gamma$ in
both domains, reaching 231--254\% at $\gamma=1.5$, within the
184--501\% range of CDP values observed for Cultural prompting, at a JSD
cost of only 0.6--3.9\% of the same-domain real-condition median. Locally, and
before boundary clipping, JSD to the original distribution grows
approximately with $(\gamma-1)^2$; moderate amplification can therefore
produce a large relative change in CDP while remaining small on the
distribution-matching scale.

% Generated by code/gamma_exaggeration.R; companion to Appendix_Controlled_Flattening.tex.
\begin{table}[t]
\centering
\small
\begin{adjustbox}{max width=\linewidth}
\begin{tabular}{llrrrrrrrr}
\toprule
Domain & $\gamma$ & CD & CDP (\%) & JSD$\to$GT & WD$\to$GT & KS$\to$GT & \shortstack{JSD / LLM\\median (\%)} & \shortstack{Mean clipped\\magnitude} & \shortstack{Max clipped\\magnitude} \\
\midrule
\multirow{3}{*}{WVS} & 1.00 & 0.0219 & 100 & 0.0000 & 0.0000 & 0.0000 & 0.00 & 0.0000 & 0.0000 \\
 & 1.25 & 0.0363 & 166 & 0.0021 & 0.0822 & 0.0317 & 0.90 & 0.0002 & 0.0075 \\
 & 1.50 & 0.0556 & 254 & 0.0091 & 0.1599 & 0.0625 & 3.89 & 0.0014 & 0.0313 \\
\midrule
\multirow{3}{*}{Big Five} & 1.00 & 0.0047 & 100 & 0.0000 & 0.0000 & 0.0000 & 0.00 & 0.0000 & 0.0000 \\
 & 1.25 & 0.0074 & 158 & 0.0003 & 0.0322 & 0.0137 & 0.15 & 0.0000 & 0.0000 \\
 & 1.50 & 0.0108 & 231 & 0.0013 & 0.0644 & 0.0275 & 0.64 & 0.0000 & 0.0020 \\
\bottomrule
\end{tabular}
\end{adjustbox}
\caption{Controlled exaggeration through $\gamma=1.5$. CDP is CD relative to human CD; GT is the original country distribution. JSD/LLM median scales JSD$\to$GT by the same-domain LLM median. Clipped magnitude is the total negative mass set to zero.}
\label{tab:exaggeration-full}
\end{table}

\section{Country-Level Cultural Divergence}
\label{app:country-cdp}
\label{app:full}

Table~\ref{tab:country-cdp} expands the block-level results to every
country--model--method cell. All entries are base-2 JS divergences. The Human
column is each human country's item-averaged divergence from the equal-weight
human centroid, not a human--LLM distance. Synthetic CD cells are classified
against the country-specific human CD shown in the Human column.
For exploratory country-level analyses, we define
$\mathrm{CDP}_{c}=\mathrm{CD}_{\mathrm{synthetic},c}/
\mathrm{CD}_{\mathrm{human},c}$, where each term uses the centroid of its
respective synthetic or human dataset. This localized ratio is distinct from
the primary domain-level CDP defined in \S\ref{sec:method-ref}.
At this resolution, Cultural is over-divergent in
33 of 36 cells and
DeepP-I is under-divergent in 33 of 36. PHub-I has the most cells inside
the human band (20 of 36, versus two for Cultural and three for DeepP-I).
These counts support the aggregate caricature/flattening pattern without
implying that it holds in every individual cell.

% AUDIT-CORRECTED country-level CD; arrows classified against the country-specific human CD.
% Requires adjustbox, multirow, booktabs, and xcolor (already loaded by tcolorbox).
\begingroup
\newcommand{\pandoracdpup}[1]{{\setlength{\fboxsep}{1pt}\colorbox{black!12}{\ensuremath{\uparrow}\,#1}}}
\newcommand{\pandoracdpdown}[1]{{\setlength{\fboxsep}{1pt}\colorbox{black!12}{\ensuremath{\downarrow}\,#1}}}
\begin{table}[!t]
\centering\small
\setlength{\tabcolsep}{4pt}
\begin{adjustbox}{max width=\textwidth}
\begin{tabular}{llccccccccccccc}
\toprule
\multirow{2}{*}{Domain} & \multirow{2}{*}{Country} & \multirow{2}{*}{\shortstack{Human\\CD}} & \multicolumn{3}{c}{Gemma-3-4B} & \multicolumn{3}{c}{Qwen3.5-9B} & \multicolumn{3}{c}{Qwen3.5-27B} & \multicolumn{3}{c}{Llama-2-13B} \\
\cmidrule(lr){4-6}\cmidrule(lr){7-9}\cmidrule(lr){10-12}\cmidrule(lr){13-15}
 & & & Cult. & PHub-I & DeepP-I & Cult. & PHub-I & DeepP-I & Cult. & PHub-I & DeepP-I & Cult. & PHub-I & DeepP-I \\
\midrule
\multirow{6}{*}{WVS} & AR & 0.012 & \pandoracdpup{0.048} & 0.014 & \pandoracdpdown{0.005} & \pandoracdpup{0.096} & 0.011 & \pandoracdpdown{0.002} & \pandoracdpup{0.053} & 0.009 & 0.008 & \pandoracdpup{0.051} & 0.007 & \pandoracdpdown{0.003} \\
 & AU & 0.026 & \pandoracdpup{0.111} & \pandoracdpdown{0.010} & \pandoracdpdown{0.002} & \pandoracdpup{0.088} & \pandoracdpdown{0.006} & \pandoracdpdown{0.003} & \pandoracdpup{0.088} & \pandoracdpdown{0.005} & \pandoracdpdown{0.004} & \pandoracdpup{0.045} & \pandoracdpdown{0.005} & \pandoracdpdown{0.003} \\
 & DE & 0.020 & 0.028 & 0.028 & \pandoracdpdown{0.003} & \pandoracdpup{0.125} & 0.019 & \pandoracdpdown{0.006} & \pandoracdpup{0.122} & 0.020 & \pandoracdpdown{0.007} & \pandoracdpup{0.031} & 0.011 & \pandoracdpdown{0.002} \\
 & IN & 0.028 & \pandoracdpup{0.058} & \pandoracdpdown{0.010} & \pandoracdpdown{0.004} & \pandoracdpup{0.100} & 0.014 & \pandoracdpdown{0.009} & \pandoracdpup{0.093} & \pandoracdpdown{0.011} & \pandoracdpdown{0.004} & \pandoracdpdown{0.013} & \pandoracdpdown{0.007} & \pandoracdpdown{0.003} \\
 & KE & 0.027 & \pandoracdpup{0.068} & \pandoracdpup{0.051} & \pandoracdpdown{0.003} & \pandoracdpup{0.138} & 0.034 & \pandoracdpdown{0.013} & \pandoracdpup{0.171} & 0.025 & \pandoracdpdown{0.005} & \pandoracdpup{0.044} & 0.022 & \pandoracdpdown{0.004} \\
 & US & 0.018 & \pandoracdpup{0.066} & 0.018 & \pandoracdpdown{0.004} & \pandoracdpup{0.059} & \pandoracdpdown{0.005} & \pandoracdpdown{0.004} & \pandoracdpup{0.049} & 0.010 & \pandoracdpdown{0.008} & \pandoracdpup{0.057} & \pandoracdpdown{0.006} & \pandoracdpdown{0.003} \\
\midrule
\multirow{3}{*}{Big Five} & AR & 0.005 & \pandoracdpup{0.018} & 0.003 & \pandoracdpdown{0.001} & \pandoracdpup{0.017} & \pandoracdpdown{0.002} & \pandoracdpdown{0.002} & \pandoracdpup{0.025} & \pandoracdpdown{0.002} & \pandoracdpdown{0.001} & \pandoracdpup{0.016} & 0.004 & 0.003 \\
 & AU & 0.005 & \pandoracdpup{0.020} & \pandoracdpdown{0.002} & \pandoracdpdown{0.002} & \pandoracdpup{0.027} & \pandoracdpdown{0.002} & \pandoracdpdown{0.002} & \pandoracdpup{0.024} & \pandoracdpdown{0.002} & \pandoracdpdown{0.001} & 0.007 & \pandoracdpdown{0.002} & \pandoracdpdown{0.002} \\
 & IN & 0.004 & \pandoracdpup{0.016} & 0.004 & \pandoracdpdown{0.002} & \pandoracdpup{0.010} & 0.004 & \pandoracdpdown{0.002} & \pandoracdpup{0.021} & 0.003 & \pandoracdpdown{0.002} & \pandoracdpup{0.014} & 0.005 & 0.003 \\
\bottomrule
\end{tabular}
\end{adjustbox}
\caption{Country-level CD. Gray $\uparrow$/$\downarrow$ marks synthetic CD above $150\%$/below $50\%$ of country-specific human CD. Cult.: Cultural; PHub-I: PersonaHub-Inspired; DeepP-I: DeepPersona-Inspired. Countries: AR, Argentina; AU, Australia; DE, Germany; IN, India; KE, Kenya; US, United States.}
\label{tab:country-cdp}
\end{table}
\endgroup

\section{Robustness Analyses}
\label{app:robustness}

\subsection{Respondent-Level Bootstrap and Country Composition}

We define the human--synthetic gap as
$\Delta=\mathrm{CD}_{\mathrm{human}}-\mathrm{CD}_{\mathrm{synthetic}}$,
so a positive value indicates lower synthetic divergence. We quantify its
uncertainty by independently resampling whole respondents within each country
and recomputing both sides ($B=2{,}000$). Resampling adds finite-sample histogram
variation, particularly to the smaller synthetic samples, and therefore tends
to increase synthetic CD and shift the gap toward zero. The estimated shifts
in Table~\ref{tab:bootstrap-loco} are negative for every PHub-I and
DeepP-I condition, making the percentile intervals conservative for
detecting flattening; for Cultural, whose gaps are negative, the same bias acts
in the anti-conservative direction but is one to two orders of magnitude
smaller than the gap itself. This shift also explains why an original point
estimate can occasionally fall outside its bootstrap interval.

The percentile intervals in Table~\ref{tab:bootstrap-loco} exclude zero for
DeepP-I in seven of eight blocks, with Llama-2-13B on Big Five as the sole
exception, and for Cultural---in the amplification direction---in all eight;
PHub-I excludes zero in five of eight. As a separate
country-composition check, we remove the same country from the human and
synthetic data and recompute the gap. The minimum leave-one-country-out gap
remains positive in all eight DeepP-I blocks, and Australia is the
worst-case omission in seven of them; the Cultural gap likewise remains
negative under every matched omission. Because Big Five contains only three
countries, its two-country results should be interpreted as a weak sensitivity
check rather than an uncertainty interval.

Table~\ref{tab:bootstrap-loco} additionally reports bias-corrected and
accelerated (BCa) intervals computed from the same replicates with
\texttt{scipy.stats.bootstrap}. On WVS, BCa corrects the downward resampling
bias: every interval contains its point estimate, and all DeepP-I and
Cultural separations are confirmed. On Big Five, however, BCa is degenerate
because the point estimate lies at an extreme of the bootstrap distribution,
causing the bias-correction factor to diverge. The BCa quantiles consequently
collapse toward the upper tail of the bootstrap distribution, yielding
zero-width or undefined intervals. These degenerate cells are marked
``---''. We therefore treat the percentile
intervals, together with the conservative direction of the resampling bias, as
primary, and BCa as corroborating evidence on WVS.

%\clearpage

\begin{table}[t]
\centering
\small
\setlength{\tabcolsep}{4pt}
\begin{adjustbox}{max width=\textwidth}
\begin{tabular}{lllrccrrl}
\toprule
Domain & Model & Method & $\Delta$ & Percentile CI & BCa CI & Shift & LOCO & Omit \\
\midrule
WVS & Gemma-3-4B & Cultural & -0.0414 & [-0.0449, -0.0381] & [-0.0448, -0.0379] & -0.0001 & -0.0276 & AU \\
 &  & PHub-I & 0.0003 & [-0.0031, 0.0023] & [-0.0019, 0.0035] & -0.0006 & -0.0042 & AU \\
 &  & DeepP-I & 0.0184 & [0.0161, 0.0193] & [0.0176, 0.0207] & -0.0007 & 0.0157 & AU \\
\cmidrule(lr){2-9}
 & Qwen3.5-9B & Cultural & -0.0790 & [-0.0823, -0.0760] & [-0.0820, -0.0754] & -0.0002 & -0.0650 & KE \\
 &  & PHub-I & 0.0071 & [0.0035, 0.0081] & [0.0059, 0.0098] & -0.0012 & 0.0035 & AU \\
 &  & DeepP-I & 0.0157 & [0.0126, 0.0163] & [0.0151, 0.0172] & -0.0011 & 0.0129 & AU \\
\cmidrule(lr){2-9}
 & Qwen3.5-27B & Cultural & -0.0742 & [-0.0788, -0.0706] & [-0.0778, -0.0698] & -0.0005 & -0.0544 & KE \\
 &  & PHub-I & 0.0087 & [0.0048, 0.0096] & [0.0078, 0.0114] & -0.0015 & 0.0049 & AU \\
 &  & DeepP-I & 0.0159 & [0.0126, 0.0163] & [0.0155, 0.0172] & -0.0014 & 0.0134 & AU \\
\cmidrule(lr){2-9}
 & Llama-2-13B & Cultural & -0.0183 & [-0.0223, -0.0162] & [-0.0206, -0.0144] & -0.0009 & -0.0103 & AR \\
 &  & PHub-I & 0.0122 & [0.0088, 0.0131] & [0.0113, 0.0142] & -0.0013 & 0.0088 & AU \\
 &  & DeepP-I & 0.0188 & [0.0161, 0.0192] & [0.0184, 0.0202] & -0.0011 & 0.0167 & AU \\
\midrule
Big Five & Gemma-3-4B & Cultural & -0.0135 & [-0.0142, -0.0122] & [-0.0147, -0.0127] & 0.0002 & -0.0055 & AU \\
 &  & PHub-I & 0.0017 & [-0.0005, 0.0017] & [0.0016, 0.0025] & -0.0010 & 0.0005 & AU \\
 &  & DeepP-I & 0.0032 & [0.0015, 0.0031] & --- & -0.0008 & 0.0023 & AR \\
\cmidrule(lr){2-9}
 & Qwen3.5-9B & Cultural & -0.0132 & [-0.0145, -0.0121] & [-0.0144, -0.0121] & -0.0001 & -0.0042 & AU \\
 &  & PHub-I & 0.0022 & [0.0002, 0.0020] & --- & -0.0010 & 0.0013 & AU \\
 &  & DeepP-I & 0.0030 & [0.0013, 0.0027] & --- & -0.0010 & 0.0021 & AU \\
\cmidrule(lr){2-9}
 & Qwen3.5-27B & Cultural & -0.0187 & [-0.0205, -0.0171] & [-0.0205, -0.0170] & -0.0001 & -0.0084 & AR \\
 &  & PHub-I & 0.0024 & [0.0004, 0.0021] & --- & -0.0011 & 0.0014 & AU \\
 &  & DeepP-I & 0.0032 & [0.0011, 0.0028] & --- & -0.0012 & 0.0024 & AU \\
\cmidrule(lr){2-9}
 & Llama-2-13B & Cultural & -0.0077 & [-0.0095, -0.0060] & [-0.0095, -0.0060] & -0.0000 & -0.0006 & AR \\
 &  & PHub-I & 0.0009 & [-0.0012, 0.0009] & --- & -0.0010 & -0.0005 & AU \\
 &  & DeepP-I & 0.0019 & [-0.0000, 0.0016] & --- & -0.0011 & 0.0011 & AU \\
\bottomrule
\end{tabular}
\end{adjustbox}
\caption{Bootstrap and leave-one-country-out (LOCO) sensitivity. $\Delta=\mathrm{CD}_{\mathrm{human}}-\mathrm{CD}_{\mathrm{synthetic}}$ (positive: flattening). CIs use $2{,}000$ paired respondent-level replicates. BCa is degenerate because the point estimate lies at an extreme of the bootstrap distribution, causing the bias-correction factor to diverge; these cells are marked ``---''. Shift is the bootstrap mean minus the point estimate; LOCO reports the worst matched-country omission. PHub-I: PersonaHub-Inspired; DeepP-I: DeepPersona-Inspired.}
\label{tab:bootstrap-loco}
\end{table}

\subsection{Country-Level Discordance and Threshold Sensitivity}

For an exploratory descriptive analysis, we mark a country condition as
discordant when its JSD to the matched human distribution is below the median
for its domain while its $\mathrm{CDP}_{c}$ is below $100t\%$.
At the prespecified descriptive cutoff $t=0.5$, 37 of 108 country conditions
meet both criteria, and all use either DeepP-I or PHub-I. Varying the
relative cutoff to $t\in\{0.4,0.5,0.6\}$ yields 31, 37, and 42 conditions,
respectively, without changing the set of represented methods. These are
overlapping country conditions, not 37 independent replications.

Table~\ref{tab:threshold-sensitivity} summarizes this cutoff sensitivity and
shows that the represented methods do not change.

\begin{table}[t]
\centering
\small
\setlength{\tabcolsep}{7pt}
\begin{tabular}{crl}
\toprule
$\mathrm{CDP}_{c}$ cutoff $t$ & Conditions & Methods represented \\
\midrule
0.4 & 31 & DeepP-I, PHub-I \\
0.5 & 37 & DeepP-I, PHub-I \\
0.6 & 42 & DeepP-I, PHub-I \\
\bottomrule
\end{tabular}
\caption{Country-level discordance across $\mathrm{CDP}_{c}$ cutoffs. Conditions are counted among the same 108 country--model--method cells. PHub-I: PersonaHub-Inspired; DeepP-I: DeepPersona-Inspired.}
\label{tab:threshold-sensitivity}
\end{table}

Table~\ref{tab:lowest-jsd} lists the four conditions with the lowest
matched-human JSD within each domain; raw JSD is not ranked across domains
because the two domains use different response scales and item structures.
All eight conditions use either DeepP-I or PHub-I, and six of the
eight have $\mathrm{CDP}_{c}$ below 50\%. Thus, strong
matched-distribution fidelity can coexist with markedly reduced cross-country
divergence even among the lowest-JSD cases.

\begin{table}[t]
\centering
\small
\setlength{\tabcolsep}{4pt}
\begin{adjustbox}{max width=\textwidth}
\begin{tabular}{llllrrr}
\toprule
Domain & Model & Country & Method & JSD & CD & $\mathrm{CDP}_{c}$ (\%) \\
\midrule
WVS & Qwen3.5-27B & United States & PHub-I & 0.050 & 0.010 & 55.1 \\
 & Qwen3.5-27B & United States & DeepP-I & 0.059 & 0.008 & 42.4 \\
 & Qwen3.5-27B & Australia & PHub-I & 0.067 & 0.005 & 19.9 \\
 & Qwen3.5-27B & Germany & PHub-I & 0.070 & 0.020 & 98.5 \\
\midrule
Big Five & Qwen3.5-9B & Australia & DeepP-I & 0.109 & 0.002 & 30.4 \\
 & Qwen3.5-27B & India & DeepP-I & 0.120 & 0.002 & 35.2 \\
 & Qwen3.5-9B & India & DeepP-I & 0.123 & 0.002 & 38.6 \\
 & Qwen3.5-9B & Argentina & DeepP-I & 0.138 & 0.002 & 35.4 \\
\bottomrule
\end{tabular}
\end{adjustbox}
\caption{Four lowest matched-human JSD conditions per domain. Raw JSD is not compared across domains; $\mathrm{CDP}_{c}$ is synthetic country-level CD relative to matched-country human CD. PHub-I: PersonaHub-Inspired; DeepP-I: DeepPersona-Inspired.}
\label{tab:lowest-jsd}
\end{table}

\subsection{Generation-Seed Sensitivity}
\label{app:seed-sensitivity}

We generate every domain--model--method condition with three independent
generation seeds (42, 1, 2) and recompute all reported metrics. The headline
patterns are unchanged in all three seeds: DeepP-I attains the lowest JSD
in the same six of eight blocks, remains under-divergent in all eight with CDP
at 14--60\%, and Cultural remains over-divergent in all
eight. Across the 48 block--metric winner comparisons, only three borderline
WD and nWD winners change with the seed (lowest raw WD in 6--7 of 8 blocks;
lowest nWD in 5--7 of 8), and the only direction-class change is
Gemma-3-4B PHub-I on WVS, which moves between 98.9\% and 103.5\%,
i.e., within the immediate vicinity of the human reference.
Table~\ref{tab:seed-sensitivity} reports CDP per
seed for every condition.

\begin{table}[t]
\centering
\small
\setlength{\tabcolsep}{4pt}
\begin{adjustbox}{max width=\textwidth}
\begin{tabular}{lllrrrr}
\toprule
 & & & \multicolumn{3}{c}{CDP (\%)} & \\
\cmidrule(lr){4-6}
Domain & Model & Method & seed 42 & seed 1 & seed 2 & Range \\
\midrule
WVS & Gemma-3-4B & Cultural & 288.9 & 276.9 & 294.8 & 17.8 \\
 &  & PHub-I & 98.9 & 103.5 & 99.6 & 4.6 \\
 &  & DeepP-I & 16.0 & 16.7 & 15.3 & 1.4 \\
\cmidrule(lr){2-7}
 & Qwen3.5-9B & Cultural & 460.5 & 448.8 & 440.2 & 20.3 \\
 &  & PHub-I & 67.8 & 70.3 & 54.5 & 15.8 \\
 &  & DeepP-I & 28.5 & 30.9 & 28.2 & 2.7 \\
\cmidrule(lr){2-7}
 & Qwen3.5-27B & Cultural & 438.5 & 446.1 & 440.9 & 7.6 \\
 &  & PHub-I & 60.5 & 61.3 & 54.7 & 6.5 \\
 &  & DeepP-I & 27.3 & 22.5 & 23.9 & 4.9 \\
\cmidrule(lr){2-7}
 & Llama-2-13B & Cultural & 183.5 & 203.0 & 198.4 & 19.5 \\
 &  & PHub-I & 44.1 & 44.2 & 44.6 & 0.5 \\
 &  & DeepP-I & 14.2 & 15.3 & 15.6 & 1.4 \\
\midrule
Big Five & Gemma-3-4B & Cultural & 389.0 & 392.1 & 410.2 & 21.3 \\
 &  & PHub-I & 64.5 & 63.4 & 64.0 & 1.1 \\
 &  & DeepP-I & 31.8 & 41.6 & 43.2 & 11.4 \\
\cmidrule(lr){2-7}
 & Qwen3.5-9B & Cultural & 383.9 & 350.8 & 377.9 & 33.2 \\
 &  & PHub-I & 53.6 & 48.1 & 53.6 & 5.5 \\
 &  & DeepP-I & 34.6 & 40.4 & 34.0 & 6.4 \\
\cmidrule(lr){2-7}
 & Qwen3.5-27B & Cultural & 500.9 & 499.6 & 486.1 & 14.7 \\
 &  & PHub-I & 47.6 & 47.9 & 56.4 & 8.7 \\
 &  & DeepP-I & 30.8 & 29.8 & 34.9 & 5.1 \\
\cmidrule(lr){2-7}
 & Llama-2-13B & Cultural & 264.8 & 279.1 & 264.0 & 15.1 \\
 &  & PHub-I & 80.9 & 72.1 & 72.3 & 8.8 \\
 &  & DeepP-I & 58.5 & 59.7 & 55.4 & 4.3 \\
\bottomrule
\end{tabular}
\end{adjustbox}
\caption{Generation-seed sensitivity. CDP is synthetic CD relative to domain-level human CD; Range is max$-$min across seeds. PHub-I: PersonaHub-Inspired; DeepP-I: DeepPersona-Inspired.}
\label{tab:seed-sensitivity}
\end{table}

\subsection{WVS Sampling-Weight Sensitivity}
\label{app:weight-sensitivity}

The primary analysis uses unweighted item-wise available cases.
Applying the WVS Wave-7 within-country post-stratification variable
\texttt{W\_WEIGHT} changes aggregate human CD from 0.021905 to 0.021730
($-0.80\%$). Across the 12 WVS model--method cells, the median absolute changes
in WD, nWD, and JSD are 1.35\%, 1.15\%, and 0.88\%, respectively, and the
largest absolute change is 4.20\%. Weighting changes none of the 12
model-by-metric winners, none of the 12 block-level CDP interpretation classes,
and none of the 72 WVS country-cell interpretation classes. The main
conclusions are therefore insensitive to this weighting choice.

% Keep the existing \section{Use of AI Assistance} immediately after this file;
% it will become Appendix G when Appendices A--F precede it.

\end{document}